%% file: paper.tex
\documentclass[journal]{IEEEtran} %
\IEEEoverridecommandlockouts
\usepackage{cite}
\usepackage{amsmath,amssymb,amsfonts}
\usepackage{algorithmic}
\usepackage[pdftex]{graphicx}
\usepackage{textcomp}
\usepackage{xcolor}

\usepackage{subcaption}
\usepackage{amssymb}
\usepackage{amsmath}
\usepackage{tikz}
\usepackage{svg}

\usepackage{hyperref}
\usepackage{url}
\def\BibTeX{{\rm B\kern-.05em{\sc i\kern-.025em b}\kern-.08em
    T\kern-.1667em\lower.7ex\hbox{E}\kern-.125emX}}
\begin{document}

\title{Towards Long Term SLAM on Thermal Imagery}

\author{%
    \IEEEauthorblockN{ Keil, C}%
    , \IEEEauthorblockN{Gupta, A.}%
    , \IEEEauthorblockN{Kaveti, P.}%
    , \IEEEauthorblockN{Singh, H.}%
    \\
    \IEEEauthorblockA{\textit{Institute of Experiential Robotics. Northeastern University, Boston MA}}\\
    \IEEEauthorblockA{\{keil.c, gupta.anik, kaveti.p, ha.singh\}@northeastern.edu}
}

\maketitle

\begin{abstract}
    Visual SLAM with thermal imagery, and other low contrast visually degraded environments such as underwater, or in areas dominated by snow and ice, remain a difficult problem for many state of the art (SOTA) algorithms. In addition to challenging front-end data association, thermal imagery presents an additional difficulty for long term relocalization and map reuse. The relative temperatures of objects in thermal imagery change dramatically from day to night. Feature descriptors typically used for relocalization in SLAM are unable to maintain consistency over these diurnal changes. We show that learned feature descriptors can be used within existing Bag of Word based localization schemes to dramatically improve place recognition across large temporal gaps in thermal imagery. In order to demonstrate the effectiveness of our trained vocabulary, we have developed a baseline SLAM system, integrating learned features and matching into a classical SLAM algorithm. Our system demonstrates good local tracking on challenging thermal imagery, and relocalization that overcomes dramatic day to night thermal appearance changes. Our code and datasets are available here: \href{https://github.com/neufieldrobotics/IRSLAM_Baseline}{https://github.com/neufieldrobotics/IRSLAM\_Baseline}
\end{abstract}

\begin{IEEEkeywords}
    Visual SLAM, Long Term SLAM, Thermal Imagery, Datasets, Learned Visual Features  
\end{IEEEkeywords}

\input{introduction}

\input{related_work}

\input{dataset}

\input{method}

\input{experiments}

\input{conclusion_and_future_work}

\bibliographystyle{plain}

\bibliography{bibdoc}

\end{document}

%% file: introduction.tex
\section{Introduction}

Simultaneous Localization and Mapping (SLAM) allows robots to track their movement, and build knowledge about their environment. Visual SLAM systems, which use camera imagery as their primary sensing modality, are pivotal for a wide variety of robotics applications, but often struggle in environments with poor visibility or significant illumination changes, such as those encountered in nocturnal or adverse weather conditions. Long-Wave Infrared (LWIR) imagery, commonly referred to as thermal imaging, emerges as a promising solution to provide visibility in dark, dust-filled, or smoke-filled environments without lighting. In addition, thermal cameras can offer significant power and weight savings when compared with LIDAR, and can offer improved visibility in autonomous driving, or other scenarios where lighting is available. Unfortunately, temperature driven appearance changes in outdoor thermal imagery that manifest over even a few hours pose unique challenges, particularly in feature extraction and localization under varied environmental conditions.

Existing feature-based methods\cite{9804793}\cite{khattak2019robust}\cite{mouats2018performance}, are notably less effective with infrared (IR) imagery. This ineffectiveness is due to reduced and inconsistent feature extraction in the short term, and inverting image gradients caused by the variations in LWIR energy across different objects in the long term. We show that inconsistent feature extraction causes the ORB\cite{orb2011} based place recognition schemes used in almost all SOTA visual SLAM systems\cite{orbslam3}\cite{qin2017vins}\cite{usenko19nfr} to be ineffective over temporal gaps of only a few hours. Leading feature-based SLAM systems, such as ORB-SLAM3 \cite{orbslam3}, encounter significant difficulties. In contrast, state-of-the-art flow-based methods like DROID-SLAM \cite{teed2021droid} provide reasonable local tracking results, but lack an easily exploitable mapping/place recognition model conducive to relocalization within diurnal LWIR datasets. Other flow-based frameworks, including VINS-FUSION\cite{qin2017vins} and Basalt\cite{usenko19nfr}, rely on BRIEF/ORB features\cite{brief} for loop closure detection or relocalization, thus faltering with IR imagery.

\begin{figure}[t]
    \includegraphics[width=\columnwidth]{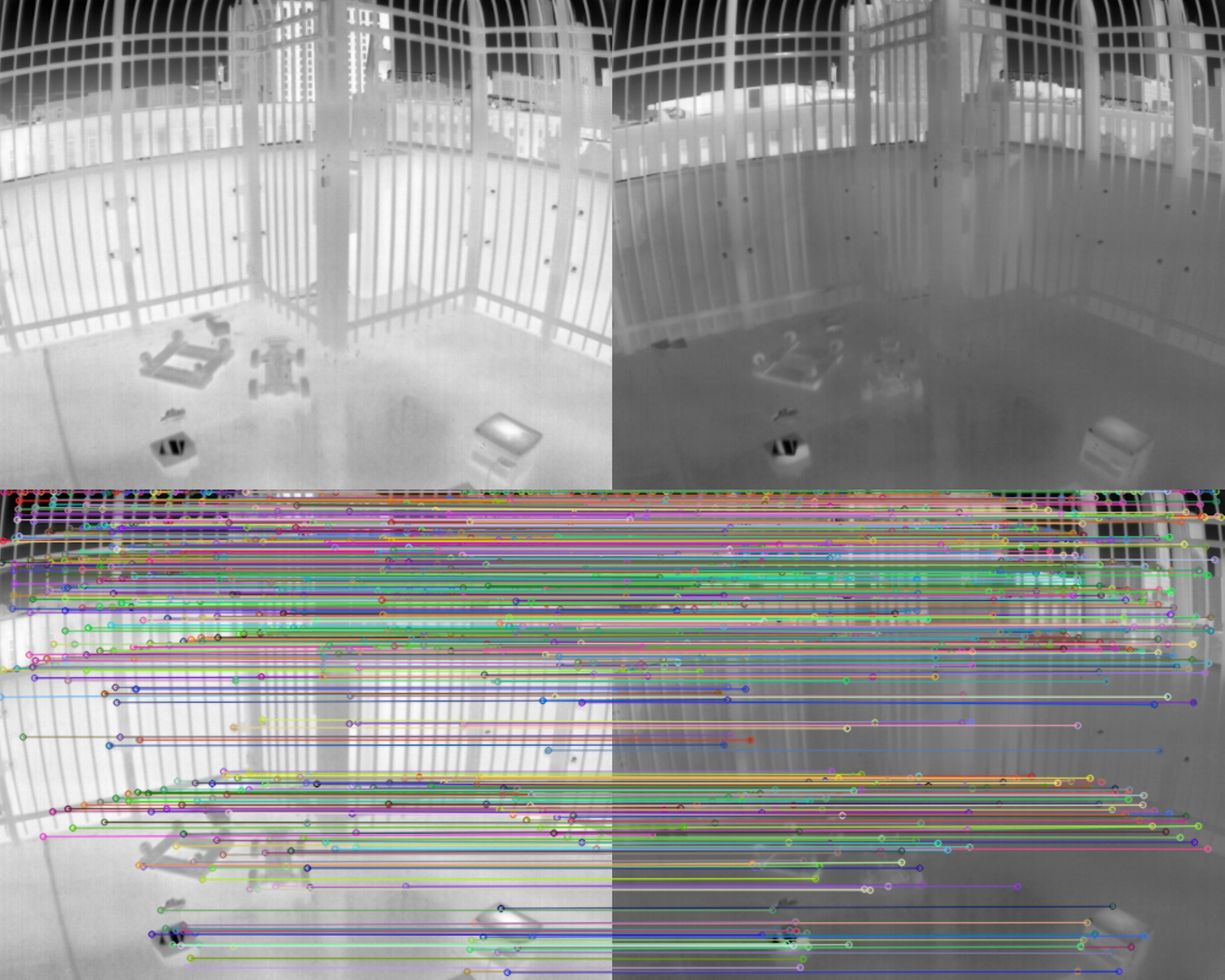}
    \caption{Long Wave Infrared (thermal) imagery poses a significant challenge for place recognition due to dramatic appearance changes over the course of a day. At the top we show a pair of images taken with a static camera approximately 12 hours apart. At the bottom we show matches that are recoverable using the Gluestick feature matching pipeline.}
    \label{fig:leader}
\end{figure}

This work endeavors to facilitate all-day autonomy for robotic systems employing LWIR cameras as their primary sensor. Our approach initially revisits classical feature extraction techniques, identifying issues with their suitability for LWIR imagery. Subsequently, we advocate for the utilization of Gluestick \cite{gluestick2023} as a learning-based method for extracting and matching features resistant to illumination changes. Lastly, we integrate this learned feature descriptor within MCSLAM \cite{mcslam} to assess its visual SLAM performance. For loop closure and relocalization testing, we develop a Bag of Words (BoW) vocabulary employing SuperPoint features from LWIR images captured at various times of the day, utilizing data gathered in urban, outdoor settings on the Northeastern Campus via handheld or vehicle-mounted cameras.

To rigorously evaluate our method, and compare against other SLAM systems, we collected comprehensive test datasets over 24-hour periods using both static and mobile IR cameras with RTK GPS ground truth. These datasets highlight the inadequacies of most existing data collections, particularly in capturing the dynamic illumination conditions inherent to outdoor environments. Our experiments indicate that our Gluestick augmented variant of MCSLAM, adeptly tracks features and achieves relocalization between day and night imagery.

Our contributions can be summarized as follows:
\begin{itemize}
    \item We present an extensive dataset collected with FLIR Boson thermal cameras. This includes a set of twenty four hour outdoor timelapses with static cameras, sequences with nearly identical camera trajectories captured in the same locations during the day and night, with GPS ground truth, and a large set of trajectories with no ground truth, which can be used for training BoW models, or for unsupervised training.    

    \item We train a BoW vocabulary using Superpoint\cite{superpoint2018} features, which we make publicly available, to show effective loop closure and Visual Place Recognition (VPR) across day-night datasets.
    
    \item We propose an effective feature based visual SLAM baseline using MCSLAM\cite{mcslam} with Superpoint\cite{superpoint2018} features and the GlueStick \cite{gluestick2023} matcher which demonstrates strong local tracking, and includes our BoW vocabulary, allowing us to save a map during the day and accurately relocalize at night.

\end{itemize}

%% file: related_work.tex
\section{Related Work}

\subsection{Feature Matching in Thermal Images}
Feature matching is a cornerstone of effective SLAM, providing the critical linkage between successive frames and across sessions. ORB \cite{orb2011}, SIFT \cite{sift2004}, and SURF \cite{bay2006surf} are foundational hand-tuned algorithms that have shaped the development of feature extraction and matching in computer vision. The performance evaluation of these features on thermal images in terms of detection, repeatability, and matching showed poor performance due to the non uniformity of noise, low contrast and flat field correction(FFC)\cite{mouats2018performance}. Learning-based features \cite{yi2016lift}\cite{simo2015discriminative} \cite{superpoint2018}\cite{tyszkiewicz2020disk}\cite{jerome2019r2d2}\cite{dusmanu2019d2} show a significant improvement in feature detection and matching, offering enhanced robustness and accuracy over classical features. However, the learned features trained on visual spectrum do not readily adapt to thermal images. Methods such as \cite{lu2021superthermal} \cite{9341716}, were specifically designed for thermal imagery. \cite{9341716} proposed augmenting the SuperPoint model with a specialized noise filter for thermal imagery and \cite{lu2021superthermal} leverages cross-spectral data to extract thermal features. These descriptors were trained and evaluated for short-term differences of a few frames to accommodate FFC. However, thermal images between day and night have dramatic differences in intensity (as shown in Fig. \ref{fig:leader}), including intensity inversions, and feature matching usually fails in these cases. In our approach, we propose to use Gluestick \cite{gluestick2023} which uses a Cross-Attention based matching scheme to robustly match images in challenging scenarios like low overlap, different lighting conditions etc.

\subsection{Thermal Simultaneous Localization and Mapping (SLAM)}
Visual SLAM is a well-explored research area with several versatile and robust methods, including sparse featured-based frameworks\cite{orbslam3}, multi-sensor systems\cite{mcslam}, and direct flow-based methods\cite{qin2017vins} \cite{usenko19nfr}. These popular classical SLAM systems require good lighting conditions and high overlap imagery to work and thus can fail in more challenging scenarios. 
Compared with visible light spectrum images, there is relatively little work in the SLAM space on thermal imagery. To overcome the limitations of thermal data, recent works have looked at fusing thermal imagery with visible spectrum imagery\cite{10372218}, LIDAR \cite{8737772} and IMU\cite{10048516}\cite{chen2023thermal}\cite{khattak2020keyframe}\cite{delaune2019thermal}. \cite{10048516} proposed an edge based feature tracking approach for improving thermal inertial odometry, while \cite{9623261}, and \cite{9804793} proposed incorporation of 14 or 16 bit thermal images for direct thermal inertial solutions. Notably, the research is mostly biased towards odometry.  

Effective place recognition and loop closure are pivotal for ensuring the consistency and reliability of SLAM systems, especially for long-term and multi-session mapping between day and night when the images look very different. Recent work on global image descriptors \cite{r2former2023}, \cite{coarse2fine2019} have shown impressive results in large real world environments, but their use has mostly been limited to offline structure from motion (SFM) applications. Bag of Words (BoW) approaches have been instrumental in achieving efficient loop closure detection with significantly lower computational requirements when compared with deep learning approaches. DBoW2\cite{GalvezTRO12}, which was first released in 2012, is still used by almost all state-of-the-art systems that employ place recognition\cite{orbslam3}\cite{qin2017vins}\cite{usenko19nfr}\cite{bad_slam}\cite{mcslam}. 
Several recent studies have blended BoW vocabularies and learned features, using SuperPoint or similar features in BoW schemes for improved robustness. These have variously incorporated novel verification metrics\cite{sp_graph2019}, examined methods for turning SuperPoint into a binary descriptor \cite{lipoLCD}, and used learned matching for verification \cite{srvio2023}. Within the domain of thermal imagery, \cite{9623261} uses the learned global descriptor approach, and \cite{9804793} uses BoW, but neither of them explores day-to-night relocalization. \cite{lee2023night} uses a generative image translation technique to approach the specific problem of day-to-night thermal relocalization, but does not show any results in a SLAM framework. 

With this context established, we organize the remainder of the paper as follows. In the next section we describe our data collection for training, analysis, and bench-marking our SLAM system. Following that we outline our methods for training a bag of words vocabulary, and SLAM integration. Finally we discuss our validation experiments and conclusions.

%% file: dataset.tex
\begin{figure}
    \centering
    \includegraphics[width=\columnwidth]{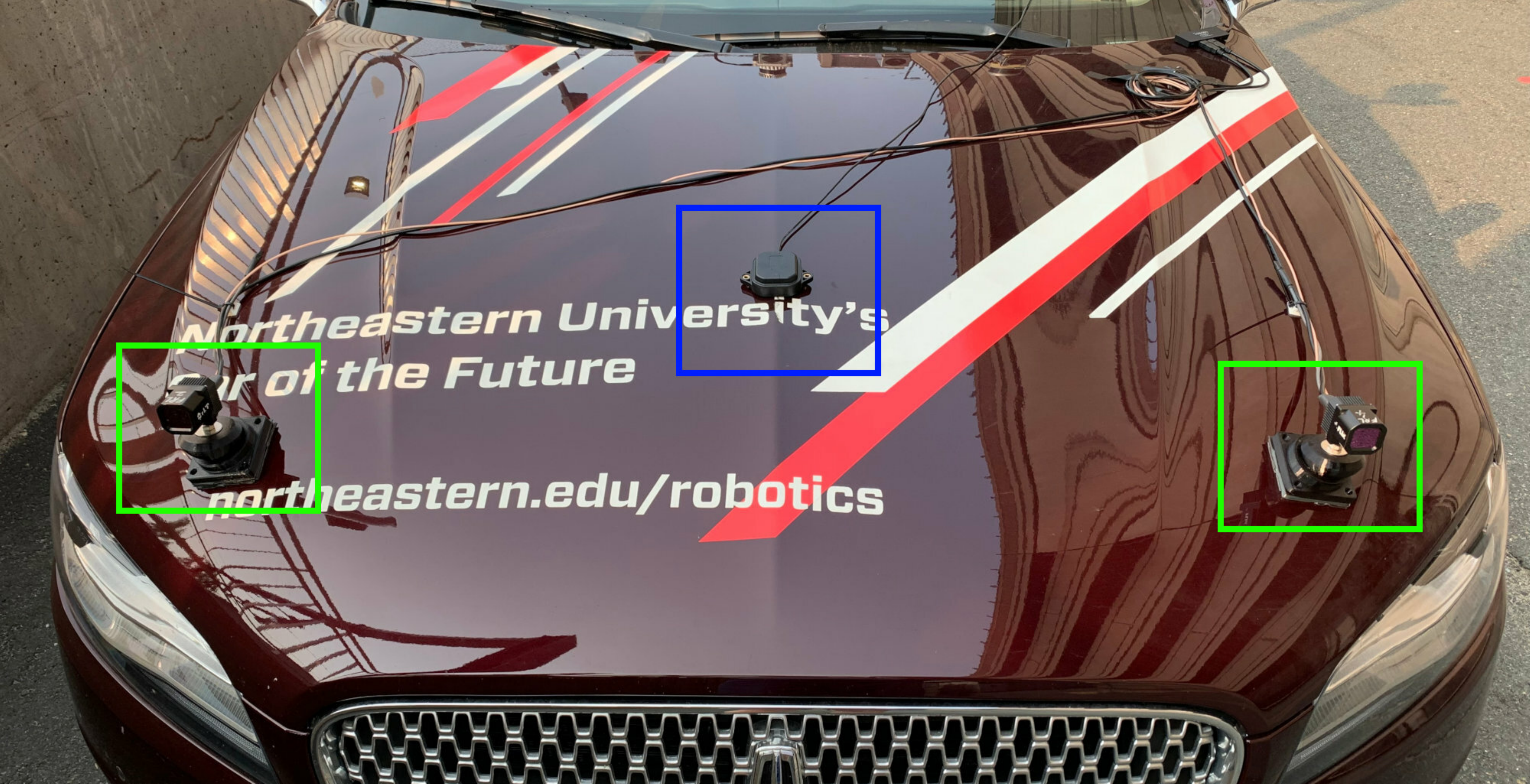}
    \caption{Data Collection setup showing two FLIR Boson ADK cameras (green), and the RTK GPS antenna (blue).}
    \label{fig:data_collection_system}
\end{figure}
\section{Dataset}
We collected a varied dataset of thermal imagery to evaluate Day-Night relocalization and SLAM performance, which we now publish. Here we describe our data collection procedure and the different formats we collected.
\subsection{Collection Types}
We collected three main types of data. First, we collected a test set of 24 hour outdoor timelapses with static cameras, in which there are very few dynamic objects. We provide ten scenes with images taken every ten minutes, showing buildings in a semi-urban environment. We use these scenes to benchmark methods where a pixel-level accurate ground truth is useful. Next, we collected a large set of monocular sequences with handheld and vehicle mounted cameras, with no ground truth trajectory information, which we use to augment our BoW training. Finally, we collected three sets of matched day and night trajectories of various sizes with stereo and side facing cameras. For these trajectories, we followed a pre-determined route during the day and then again at night, taking care to be as consistent as possible between attempts. We collect at least two loops at each time so that place recognition evaluations can be carried out for day-to-day/night-to-night and also for day-to-night, and visa versa. The features of these datasets are summarized in table \ref{Tab:evaluation_traj}. For convenience we refer to the sequences by the names given in table \ref{Tab:evaluation_traj}. For these sets we provide ground truth position data with RTK GPS, making for an easily verifiable set for benchmarking day-to-night and night-to-day loop closure. Images from the "KRI" sequence are shown in figure \ref{fig:feature_compare_qualitative}. In our subsequent evaluation we use the "KRI" loop as our test set and retain the others for training.

\begin{table}[t]
\begin{tabular*}{\columnwidth}{@{\extracolsep{\fill}}|l|p{6.5cm}|}

\hline
Name                               & Description                                                                   \\ \hline
Garage Roof  & Small loops taken on top of a parking garage, with few dynamic objects.    \\ \hline
Carter Field & Medium sized loops on a public road in Boston, with mobile cars and pedestrians. \\ \hline
KRI & Larger loops at the Kostas Research Institute, with few dynamic objects.                       \\ \hline
\end{tabular*}%
\caption{SLAM Evaluation Datasets.}
\label{Tab:evaluation_traj}
\end{table}

\subsection{Hardware Setup}
All of our images were captured using a FLIR Boson ADK cameras, with a resolution of 512x640, and a horizontal field of view of 75 degrees. In the case of the paired day-night trajectories, we use a stereo pair with a baseline of 1.1m mounted on a Lincoln MKZ car. The stereo pair and RTK GPS antenna placement can be see in figure \ref{fig:data_collection_system}. We capture frames at 30fps for paired datasets, and at 60fps in the monocular trajectories. 

\subsection{Calibration}
Traditional camera calibration targets show little to no contrast between white and black regions when viewed with a thermal camera. We use a wooden calibration target with a checkerboard pattern formed from copper tape. When the calibration target is heated with an external source the wood appears white, and the copper squares, which reflect the ambient environment, appear dark, providing enough contrast for calibration. We use Kalibr\cite{furgale2013unified} to estimate the intrinsic coefficients for all cameras and the extrinsic parameters for our stereo pair.

%% file: method.tex
\section{Method}
\subsection{Image Preprocessing}
Raw IR images output by the Boson camera have very poor contrast. We apply Contrast Limited Adaptive Histogram Equalization (Clahe)\cite{1002415}, which increases contrast thereby increased keypoint extraction, at the cost of amplifying noise. There is a trade-off between increasing the number of keypoints extracted, and increasing the the noise in the location of the extracted keypoints, especially for ORB features. We empirically determined a set of CLAHE parameters that perform well on our imagery in the context of SLAM. Note that noise in the IR images is characterized by a grid like overlay called fixed pattern noise\cite{9341716}. An example of processed images can be seen in figure \ref{fig:calhe_ims}.
\input{feature_comparison_figure}
\subsection{Gluestick}
\begin{figure}
    \centering
    \includegraphics[width=\columnwidth]{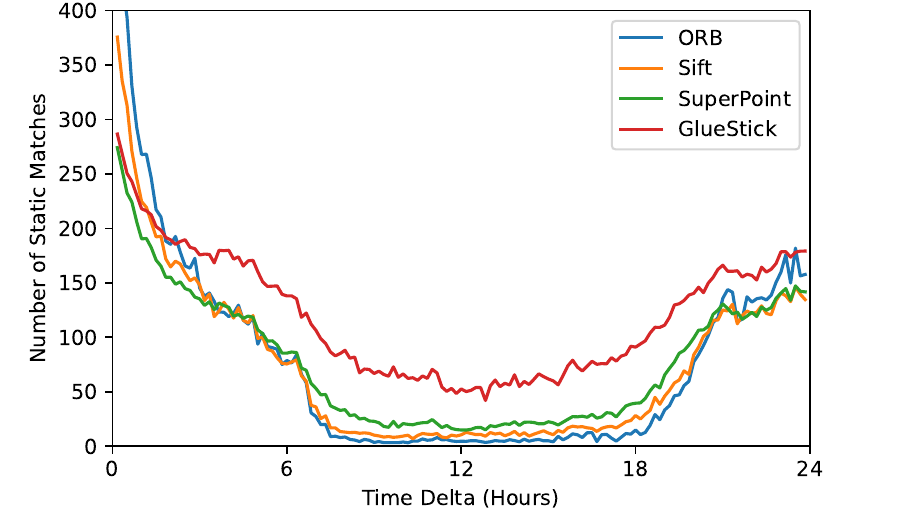}
    \caption{Mean number of matches at each time-step with error of less than three pixels on our timelapse dataset. There are 10 static outdoor scenes with images recorded every 10 minutes over a 24 hour period. For ORB, Sift, and SuperPoint we use a brute force matcher.}
    \label{fig:timelapse}
\end{figure}

Our use of Gluestick\cite{gluestick2023} is motivated by the qualitative analysis show in figure \ref{fig:feature_compare_qualitative}. We show a pair of images captured at the same time during the day with a small camera translation, and a third image from the same location captured at night, with feature matches for ORB, SIFT, SuperPoint, and Gluestick. We can clearly see that the learned features perform much better in both the day-day and day-night scenarios. Gluestick is notably better at matching features in the foreground, which is important for accurately measuring camera translation. Figure \ref{fig:timelapse} shows a quantitative comparison for the same features using our timelapse dataset to evaluate matches across a twenty four hour period. We again see that Gluestick significantly outperforms the other methods. ORB and Sift features struggle particularly with day to night matching, because they rely on the corner gradient orientation to for orientation invariance. Changing temperature gradients over time result in completely different feature orientations and descriptors, or in features dropping below detection thresholds entirely. The learned features do not explicitly use a feature orientation, have significantly greater numerical complexity, and are trained to be robust to noise, resulting in better performance.

With the above analyis in mind, we use Gluestick, to generate and match SuperPoint features for training our BoW vocabularies, and also for front-end SLAM data association. Gluestick notably also matches lines from image pairs, which are represented by the SuperPoint descriptors at the line endpoints. We observed that the matched lines are often qualitatively correct, but the locations of endpoints are obviously variable from frame to frame. We use the matched line features for place recognition, but do not use them to estimate camera motion. Initially we attempted to use Superpoint features alone, matched using classical techniques, because this combination can show reasonable results on image pairs. We modified ORB SLAM3 to use SuperPoint features, but generally found that this system had poor performance and would lose tracking frequently. Gluestick produces better matches, and introduces far fewer incorrect matches, yielding much better performance in our final SLAM pipeline. In our experiments we achieve good results using the pretrained weights for Gluestick. In the future, retraining for IR imagery is possible with a more sophisticated data collection scheme to generate sufficiently diverse IR imagery with depth and pose ground truth.

\subsection{Place Recognition}
We extend DBoW2\cite{GalvezTRO12} to build a vocabulary for Superpoint features extracted from IR images, with the aim of achieving good day-night/night-day place recognition. To train the vocabulary we match pairs of images from sequences in our training sets using Gluestick and then use only the features that are successfully matched across the pairs as training input for the vocabulary. We train with approximately 38,000 pairs of images from a mix of standalone sequences, and paired day-night sequences captured across several months. Omitting the paired trajectories, using only day or only night images resulted in poor temporal generalization, suggesting that online vocabulary building methods would not work well here. Note that we do not explicitly match across day and night images, and only use sequential image pairs from within the same sequence. We are able to achieve compelling results without handling day to night matches in a principled way. In the future we can improve on this by using GPS information to select day-night pairs from a dataset, and explicitly train a vocabulary with day to night matches. Our final vocabulary has five levels and a branching factor of 10, which is 10x smaller than the vocabulary used by ORB-SLAM. Larger vocabularies exhibit poor performance on day to night matches in our testing, likely as a result of over-fitting.

The main advantage of using binary features for place recognition in a BoW scheme is that they are extremely fast to compute. In our implementation, the L2 distance computation for comparing Superpoint's floating point valued features takes roughly 80 times as long to compute as the Hamming distance used for brief features. However, even with this performance constraint, the single threaded DBoW2 implementation running on a modern laptop can add Superpoint features to a large database and search for match candidates at over 50 frames per second, which is easily fast enough for real world SLAM loop closure. In a typical BoW scheme the majority of the computation time for adding an image or querying a database is dedicated to computing the BoW feature vector for an image. Once a BoW feature vector has been computed, the actual lookup time to search a database is essentially the same regardless of the original feature type (assuming a similar number of features per image, and distribution of those features in a database) because the comparisons are on words in the vocabulary, which have the same datatype for ORB or SuperPoint, or any other feature.

\subsection{SLAM Pipeline}
We chose to use Multi Camera SLAM \cite{mcslam}, MCSLAM as the basis of our SLAM system. MCSLAM is a flexible feature based framework, partially based on ORB-SLAM, which supports camera setups beyond the typical stereo-pairs supported by most popular SLAM systems. In this work we limit our analysis to stereo imagery, however due to the low resolution of IR cameras and general difficulty of working in IR imagery, this choice also allows us to consider arrays of cameras with overlapping and non overlapping fields of view for a more robust system in the future, or examine combined arrays of RGB and IR cameras.

In our implementation we replace ORB features with Superpoint descriptors, and improve most of the matching framework with Gluestick. The open source implementation of Gluestick works as an end to end pipeline, taking a pair of images and outputting a prediction which contains all of the extracted features and matches. We made some minor changes to the model so that instead of working in one step, we can individually extract features for each image, and then separately match those features with as many other sets of features as needed. In the MCSLAM workflow, images that are taken at the same time are referred to as a multi-camera frame. In our adaptation, for each multi-camera frame (always stereo pair in this work) we extract features for each image and then use Gluestick to find matches across the cameras. These matches become 3D landmarks, while unmatched features become monocular landmarks. We then associate the current multi-camera frame with the previous one by matching each camera's current image features with the previous set of features for that camera. The resulting set of associations between the current and previous frames can then be used by the MCSLAM backend to establish the relationship to the 3D map. 
MC SLAM currently supports a BoW loop closure mechanism based on DBoW2 and DLoopDetector\cite{GalvezTRO12}, which allows us to use our SuperPoint vocabulary to achieve relocalization. We use the stereo features matched in each frame, rather than the full set, to help limit the search to stronger features.

%% file: feature_comparison_figure.tex
\begin{figure*}[!t]
    \begin{subfigure}{0.75\textwidth}
        \centering
        \includegraphics[width=\textwidth]{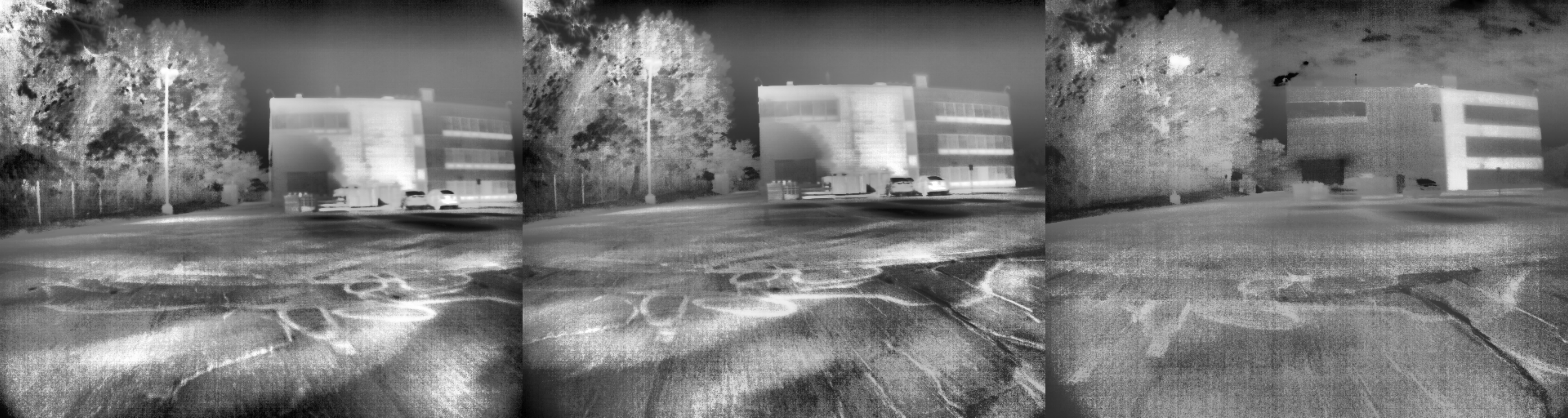}
        \caption{Images with CLAHE preprocessing}
        \label{fig:calhe_ims}
    \end{subfigure}
    \centering
    \begin{subfigure}{0.75\textwidth}
        \centering
        \includegraphics[width=\textwidth]{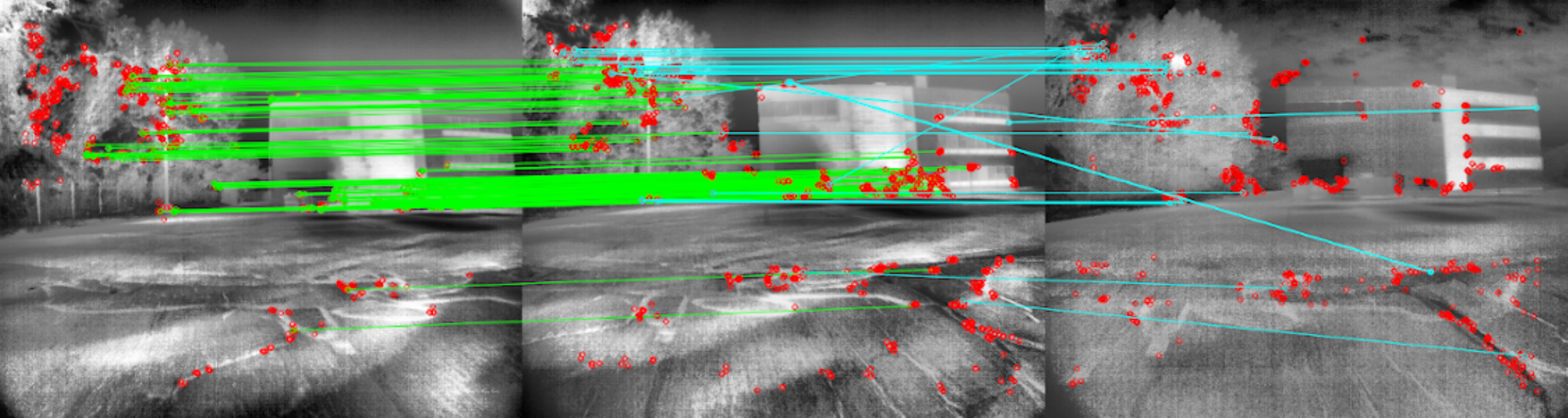}
        \caption{ORB features}
        \label{fig:orb_features}
    \end{subfigure}
    \begin{subfigure}{0.75\textwidth}
        \centering
        \includegraphics[width=\textwidth]{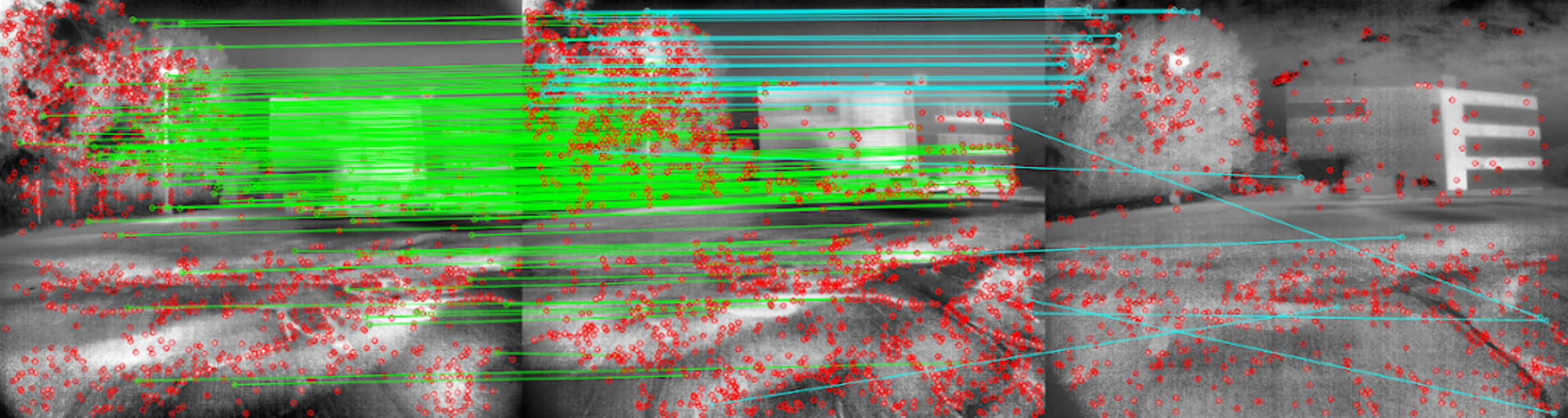}
        \caption{Sift features}
        \label{fig:sift_features}
    \end{subfigure}
    \begin{subfigure}{0.75\textwidth}
        \centering
        \includegraphics[width=\textwidth]{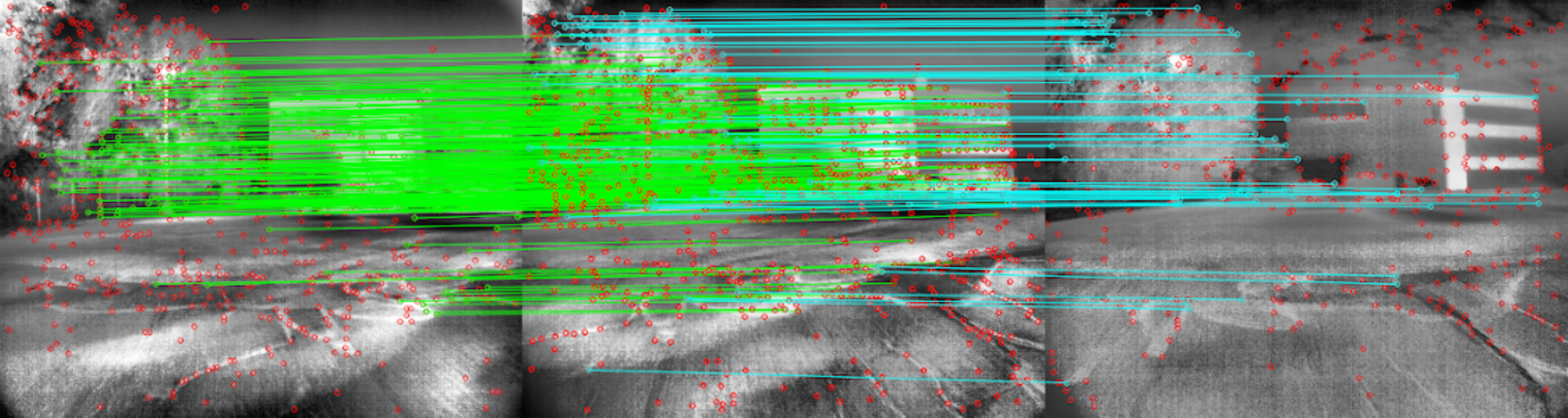}
        \caption{SuperPoint features with brute force matcher}
        \label{fig:sp_features}
    \end{subfigure}
    \begin{subfigure}{0.75\textwidth}
        \centering
        \includegraphics[width=\textwidth]{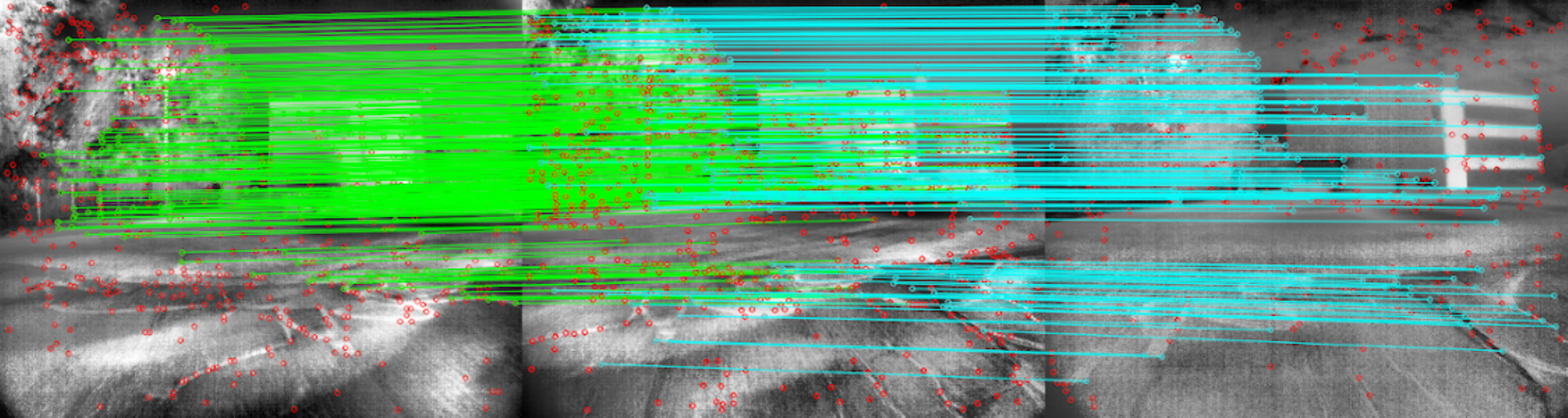}
        \caption{SuperPoint features with Gluestick matcher}
        \label{fig:gs_features}
    \end{subfigure}
    
    \caption{We show a qualitative example of the number of matched features across images taken during the day (left pair) and then the same scene at night (right pair). At the top we show preprocessed images, below that we show ORB, Sift, SuperPoint and SP+Gluestick. Features are matched with brute force matching and are filtered for geometric consistency using a RANSAC fundamental matrix estimation. SP features with the Gluestick matcher outperform all other methods and are notable better at matching features in the foreground, which is important for parallax in feature based pose estimation.}
    \label{fig:feature_compare_qualitative}
\end{figure*}

%% file: experiments.tex
\section{Experiments}

\subsection{Place Recognition}
Here we demonstrate our place recognition system in isolation from the full slam system in order to systematically demonstrate it's effectiveness. First, we test loop closure with a minimal temporal gap on the KRI test dataset. Running a separate experiment for day and night, we split the trajectories, using the first loop to build a BoW database and then search that database with queries from the second loop. The descriptors used to build and query the database are matched features from sequential frames in the sequence. We test the database using 100 uniformly distributed images and take the best scoring candidate as a loop closure, after rejecting false positives by requiring a minimum number of matches be RANSAC inliers of a fundamental matrix estimation. We set the false positive rejection threshold to the minimum required to eliminate all false positives, as is typically done in relocalization experiments\cite{GalvezTRO12}. We compare against ORB features using the vocabulary distributed with ORB-SLAM, and ground truth the experiment using GPS. We are able to detect an image with strong overlap in 100\% of the cases for both day and night, with ORB performing nearly as well. See the results in table \ref{tab:bow_kri_label}.

Next, to demonstrate results for a significant temporal window, we perform the same analysis but search for 100 frames from the KRI day set against a database of KRI night images, and visa versa. We are able to show 91\% and 93\% success respectively, with no false positives. The same experiment with ORB features shows only 12\% and 10\% respectively. See table \ref{tab:bow_kri_label} for details. In general, ORB features will not work for place recognition across a significant temporal gap, and by extension, map reuse, for thermal imagery. Gluestick shows promise.

\begin{table}[]
\centering
\begin{tabular}{lcc}
            & Ours  & Orb Vocabulary \\ \hline
Day-Day     & 100\% & 100\%           \\ \hline
Night-Night & 100\% & 97\%           \\ \hline
Search Day DB with Night Images  & 93\%  & 12\%           \\ \hline
Search Night DB with Day Images   & 91\%  & 10\%           \\ \hline
\end{tabular}
\caption{Recall at 100\% precision for our IR SuperPoint vocabulary and an ORB vocabulary, using our KRI test dataset.}
\label{tab:bow_kri_label}
\end{table}

\subsection{SLAM}
Actually re-localizing relative to a previous map is significantly more challenging than finding loop closure candidates from a set of images. There is significant added complexity from a software engineering standpoint, and the SLAM system must be able to recover an accurate 3D pose from triangulated matches. With this in mind, we test our augmented MCSLAM in two ways. First we analyze our front end tracking on IR trajectories. Second, we examine relocalization across the day-night temporal gap. In all cases we use the KRI trajectories as the basis of our comparison, because we used the other trajectories to train our BoW model.
\begin{figure}
    \centering
    \includegraphics[width=\columnwidth]{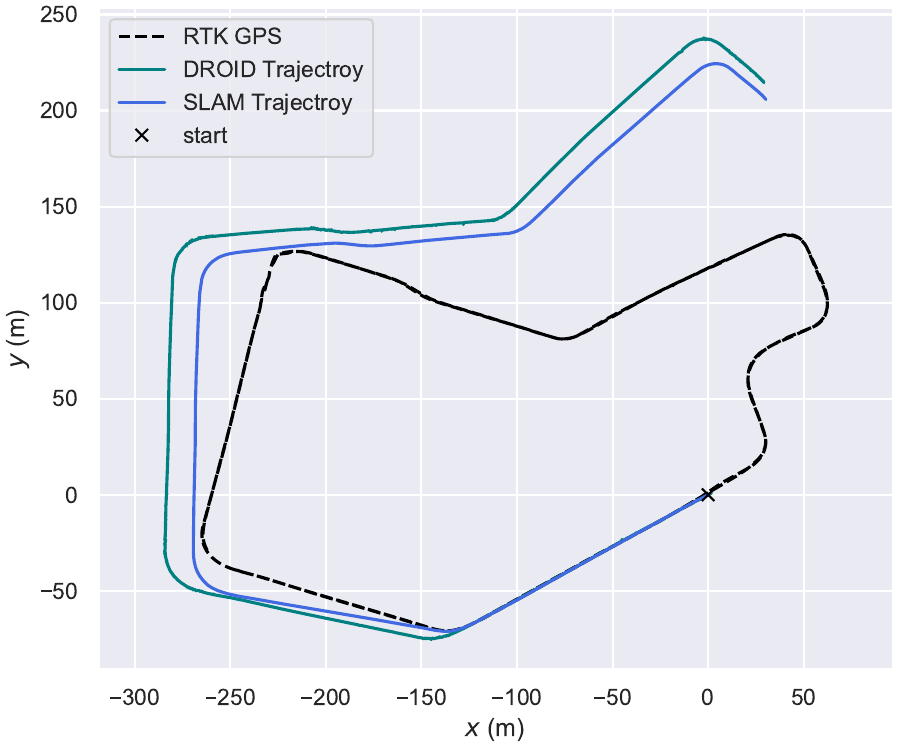}
    \caption{Here we show our Day trajectory for the KRI dataset, along with the gps ground track. ORBSLAM and ORB-MCSLAM are only able to track for small sections of the map. We show a comparison with Droid SLAM, even though it is not directly relevant to our work because it shows similar tracking performance. There is significant accumulated drift, but scale and qualitative features are correct. The trajectory ends due to a very difficult low texture region. Note that Droid SLAM is able to track longer but we have stopped it at the same point for visual clarity.}
    \label{fig:SLAM-traj}
\end{figure}
\subsubsection{Front End Tracking}
Figure \ref{fig:SLAM-traj} shows a qualitative comparison of our results on the KRI day dataset, relative to the GPS ground truth. We do not show comparisons against any feature based SLAM methods, because MCSLAM\cite{mcslam} and ORB SLAM\cite{orbslam3} are both only able to track in stereo for short periods. We observed that ORB features can be matched from frame to frame, but over a sequence noisy keypoint extraction makes it difficult to build a 3D map. The results from the Day trajectory show that we are able to generate reasonably accurate trajectories, implying similarly accurate maps. We are not able to maintain tracking through the entire trajectory due to one difficult point where there are minimal features near enough to the camera for tracking, highlighting the difficulty of using IR data for SLAM. 

\subsubsection{Relocalization}

\begin{figure}
    \centering
    \includegraphics[width=\columnwidth]{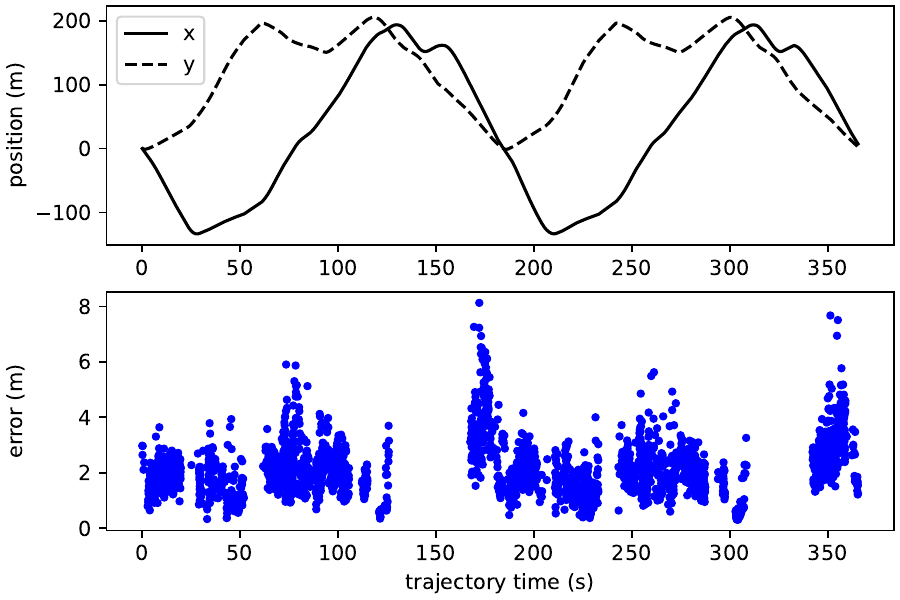}
    \caption{Here we show estimated relocalization error computed from the KRI night sequence relocalized into the KRI day map. The upper plot shows the ground truth trajectory x and y positions (east-west and north-south). The bottom plot shows the magnitude of the estimated error at corresponding points in the trajectory. Note that the trajectory contains two approximately identical loops, and that the large gaps in the error plot correspond to the section of the trajectory that is not included in the day map. We can see that the trend of relicalization error is consistent across the two cycles. The same results are shown overlain on the map in figure \ref{fig:reloc_overall}}
    \label{fig:reloc_error}
\end{figure}

To demonstrate relocalization within a map, we use the saved map from our daytime trajectory shown in figure \ref{fig:SLAM-traj}, and attempt to relocalize at every time step in the night trajectory. Relocalization for a single stereo frame is achieved by searching a DBoW2 database for similar images, removing false positive with an island identification procedure similar to the implementation in DLoopDetector\cite{GalvezTRO12}, matching the features from the map images with the query images using Gluestick, associating those image features with 3D points in the saved map, and then computing a perspective n-point solution with GTSAM\cite{gtsam} to estimate the camera's pose in the map. We threshold a minimum number of RANSAC inliers to reject low quality estimations, and reject a small number relocalization poses that are more than 10m from pose of the map keyframe returned by our BoW search. Our map has a noticeable drift over time, so we estimate the relocalization error by locally aligning the map to its GPS ground truth, applying the alignment transformation to our relocalized position, and computing the error between the transformed relocalized position and the GPS ground truth for the relocalization frame. The local alignment is based on the Umeyama alignment\cite{allignment} implementation in EVO\cite{grupp2017evo}. The resulting error is show in figure \ref{fig:reloc_error}. In the majority of cases we are able to relocalize with less than 3m of error. We can achieve single shot relocalization over most of the map, with low error relative to the size of the map, and make the general observation that larger relocalization error is not attributed to incorrect BoW candidate selection, but to noisy pose estimation resulting from the majority of visual features being on distant objects in some areas of the map.

\begin{figure}
    \centering
    \includegraphics[width=\columnwidth]{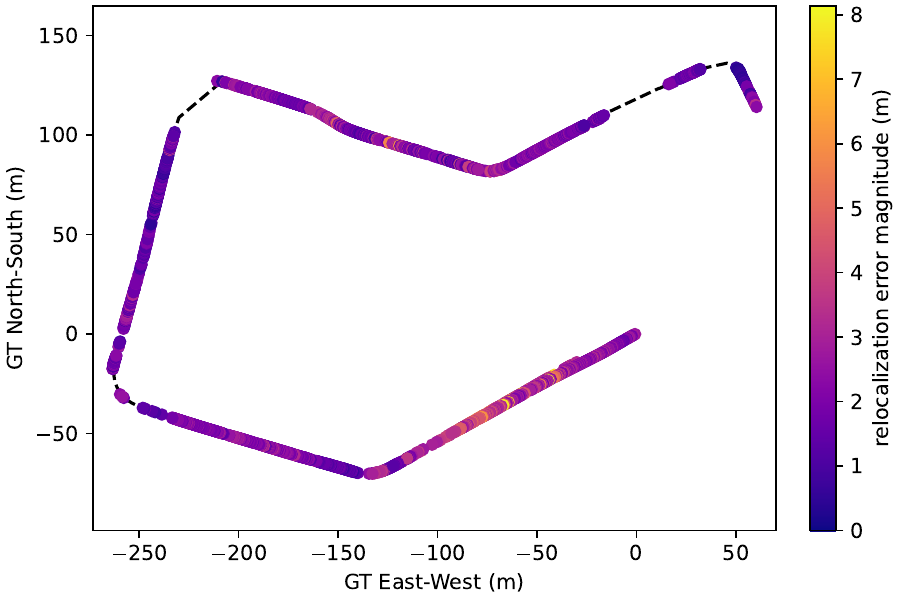}
    \caption{Here we see the error magnitude heatmap of the the night trajectory relocalized into the day map. At certain points in the trajectory, the average error spikes. This generally happens because in these regions the strongest features tend to be on distant structures.}
    \label{fig:reloc_overall}
\end{figure}

%% file: conclusion_and_future_work.tex
\section{Conclusion and Future Work}
We have shown that challenging loop closure and relocalization, enabling map reuse, is possible within long term IR datasets. We are able to achieve this using a BoW system, making it suitable for relatively simple incorporation into existing SLAM systems. Our baseline SLAM system is able to generate maps using Gluestick for data association, and outperforms feature based SLAM systems that use binary descriptors. One avenue for future work would be to better optimize the system for memory use and speed. Gluestick is not a perfect drop in replacement for the efficient matching scheme used in MCSLAM, or other feature based methods. Improvements and optimizations could be made with regards to matching across more than two frames, and matching between cameras with known extrinsic parameters. It is also worth noting that our datasets are hardly comprehensive. In the future we will be looking at collecting larger, more diverse datasets that include other sensing modalities for comparison, enabling us to retrain or fine-tune feature extraction and matching, build better vocabularies, and conduct more thorough analysis.